\documentclass[sigconf]{acmart}

\usepackage{bm}
\usepackage{cleveref}
\usepackage{multirow}
\usepackage{booktabs}
\usepackage[table]{xcolor}
\definecolor{highlightgray}{gray}{0.9}
\usepackage{graphicx}

\usepackage{balance}

\AtBeginDocument{%
  }

\copyrightyear{2025}
\acmYear{2025}
\setcopyright{acmlicensed}
\acmConference[MM '25] {Proceedings of the 33rd ACM International Conference on Multimedia}{October 27--31, 2025}{Dublin, Ireland.}
\acmBooktitle{Proceedings of the 33rd ACM International Conference on Multimedia (MM '25), October 27--31, 2025, Dublin, Ireland}
\acmISBN{979-8-4007-2035-2/2025/10}
\acmDOI{10.1145/3746027.3755225}

\begin{document}

\title{Multi-Agent Amodal Completion: Direct Synthesis with Fine-Grained Semantic Guidance}


\author{Hongxing Fan}
\affiliation{%
\department{School of Computer Science and Engineering}
  \institution{Beihang University}
  \city{Beijing}
  \country{China}}
\email{fanhongxing@buaa.edu.cn}
\orcid{0009-0003-4525-8777}

\author{Lipeng Wang}
\affiliation{%
\department{School of Software}
  \institution{Beihang University}
  \city{Beijing}
  \country{China}}
\email{wanglipeng@buaa.edu.cn}
\orcid{0009-0003-9100-7899}

\author{Haohua Chen}
\affiliation{%
\department{School of Software}
  \institution{Beihang University}
  \city{Beijing}
  \country{China}}
\email{robertchen245@buaa.edu.cn}
\orcid{0009-0005-4631-0172}

\author{Zehuan Huang}
\authornote{Project Lead}
\affiliation{%
\department{School of Software}
  \institution{Beihang University}
  \city{Beijing}
  \country{China}}
\email{huangzehuan@buaa.edu.cn}
\orcid{0009-0002-1883-0777}

\author{Jiangtao Wu}
\affiliation{%
\department{School of Software}
  \institution{Beihang University}
  \city{Beijing}
  \country{China}}
\email{wujiangtao@buaa.edu.cn}
\orcid{0009-0003-9288-3823}

\author{Lu Sheng}
\authornote{Corresponding Author.}
\affiliation{%
\department{School of Software}
  \institution{Beihang University}
  \city{Beijing}
  \country{China}}
\email{lsheng@buaa.edu.cn}
\orcid{0000-0002-8525-9163}

\renewcommand{\shortauthors}{Hongxing Fan et al.}


\begin{abstract}
Amodal completion, generating invisible parts of occluded objects, is vital for applications like image editing and AR. Prior methods face challenges with data needs, generalization, or error accumulation in progressive pipelines. We propose a Collaborative Multi-Agent Reasoning Framework based on upfront collaborative reasoning to overcome these issues. Our framework uses multiple agents to collaboratively analyze occlusion relationships and determine necessary boundary expansion, yielding a precise mask for inpainting. Concurrently, an agent generates fine-grained textual descriptions, enabling Fine-Grained Semantic Guidance. This ensures accurate object synthesis and prevents the regeneration of occluders or other unwanted elements, especially within large inpainting areas. Furthermore, our method directly produces layered RGBA outputs guided by visible masks and attention maps from a Diffusion Transformer, eliminating extra segmentation. Extensive evaluations demonstrate our framework achieves state-of-the-art visual quality.
\end{abstract}




\begin{teaserfigure}
\centering
  \includegraphics[width=1\textwidth]{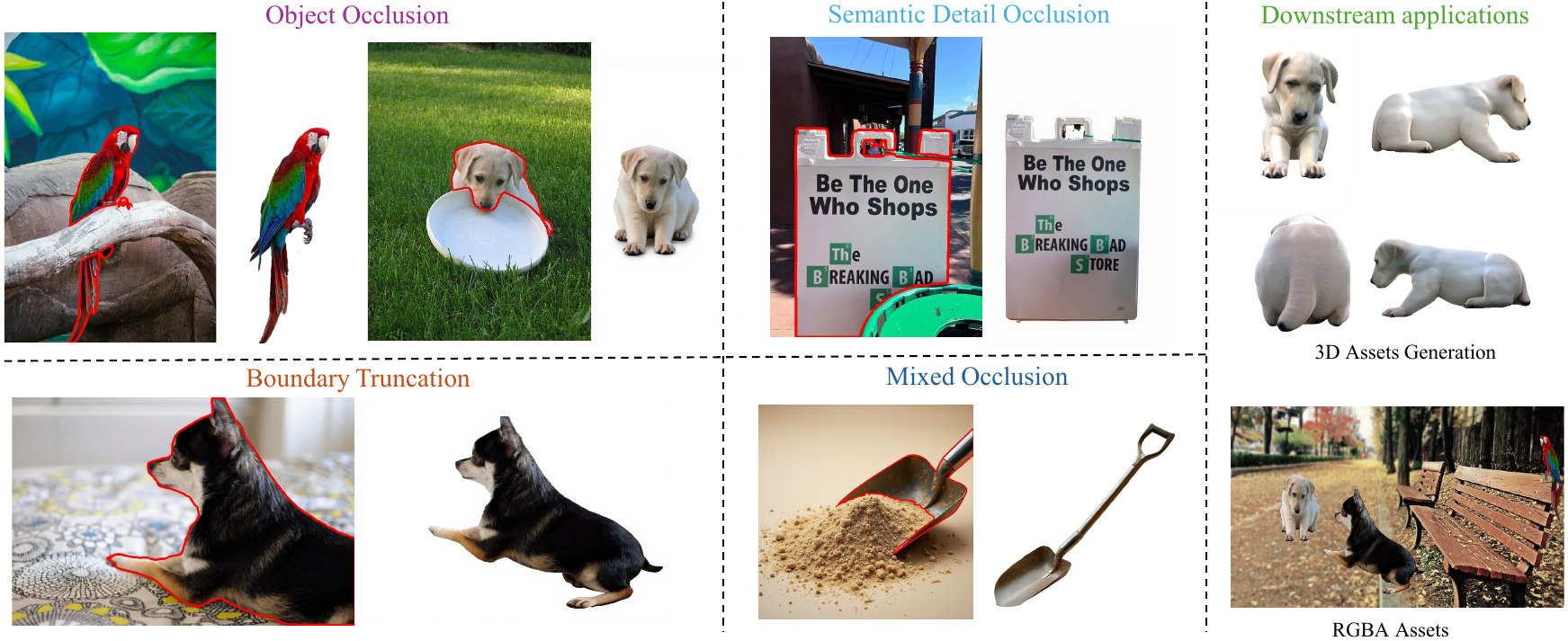}
  \caption{Our framework robustly handles diverse amodal completion challenges, from Object Occlusion and Boundary Truncation to Mixed and Semantic Detail Occlusion. The generated RGBA outputs directly enable diverse downstream tasks.}
  \label{fig:teaser}
  \Description{A composite image demonstrating the framework's capabilities across four categories, plus downstream applications. The top-left quadrant, labeled "Object Occlusion," shows a red macaw partially hidden by a tree branch and a puppy partially hidden by a white frisbee, with the framework generating the complete, unobstructed versions of both animals. The bottom-left, "Boundary Truncation," shows a dog cut off by the image edge, with the framework generating its complete body. The top-right, "Semantic Detail Occlusion," shows a sign with text partially obscured, and the framework restores the full text "Be The One Who Shops. The BREAKING BAD STORE." The bottom-right, "Mixed Occlusion," shows a shovel partially buried in sand, and the framework reveals the complete shovel head. The final column, labeled "Downstream applications," shows how the generated assets can be used, with examples labeled "3D Assets Generation" and "RGBA Assets."}
\end{teaserfigure}


\maketitle

\section{Introduction}

Amodal completion is the task of inferring and generating invisible parts of a partially occluded object. It is a critical capability for various applications, including image editing, 3D modeling, and augmented reality\cite{van1999investigating,yun2018temporal}. By accurately hallucinating the occluded region, the completed object can serve as a valuable digital asset for downstream applications.

Previous approaches to amodal completion are typically training-based\cite{bowen2021oconet,zhan2020self,ehsani2018segan,yan2019visualizing,zhou2021human,ling2020variational,zhan2024amodal,ozguroglu2024pix2gestalt}, often employing either two-step pipelines (mask prediction then inpainting) or leveraging deep generative models\cite{goodfellow2014generative,kingma2013auto,ho2020denoising} to learn object priors from data. However, these training-dependent methods generally suffer from reliance on large task-specific datasets and struggle with generalization. Consequently, they face difficulties with severely occluded objects and exhibit limited robustness in out-of-distribution scenarios.

\begin{figure}[t]
    \centering
    \includegraphics[width=1\columnwidth]{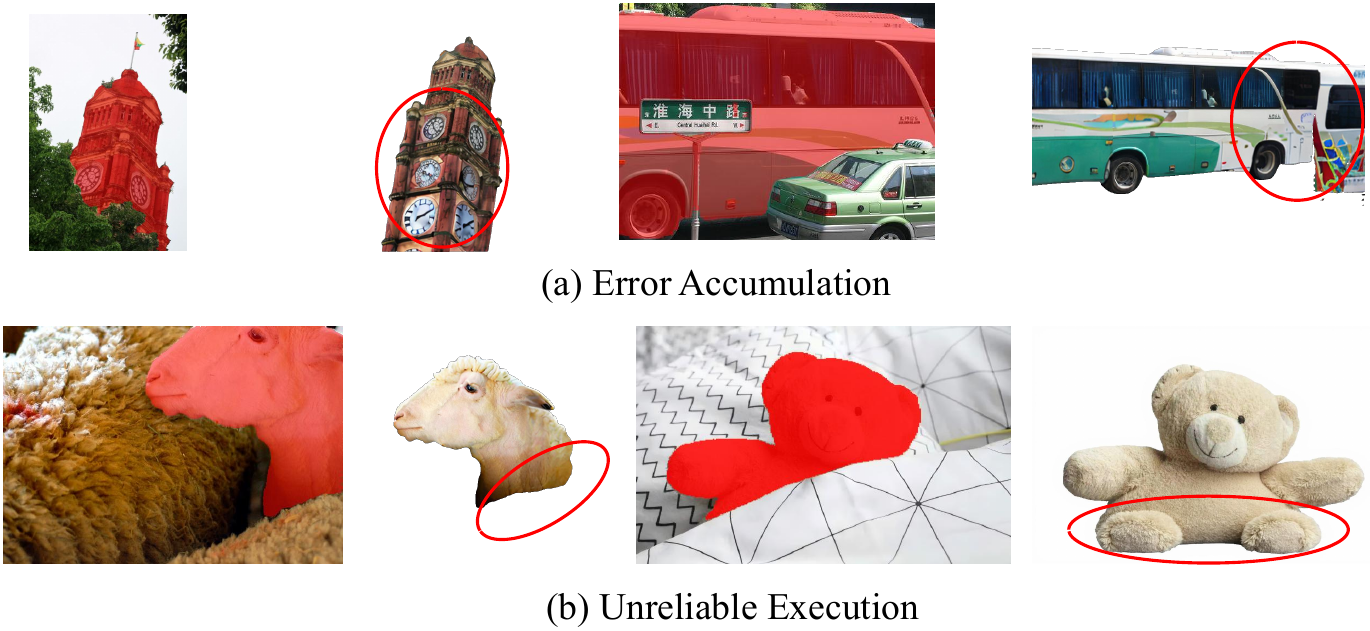}
    \caption{Common failure modes of prior progressive methods. (a) Error Accumulation: Early errors propagate during iteration, causing inconsistencies/artifacts. (b) Unreliable Execution: Progressive processes can fail prematurely or inaccurately. Our upfront reasoning avoids these iterative errors.}
    \label{fig:progressive failure examples}
    \Description{A figure illustrating two common failure modes of prior progressive methods, split into two rows. Row (a), labeled "Error Accumulation," shows two examples. The first example shows a red tower that is incorrectly extended downwards with multiple extra clock faces. The second shows a bus incorrectly extended with distorted and misplaced windows. Row (b), labeled "Unreliable Execution," also shows two examples. The first shows a sheep's head truncated by the image boundary, where the amodal completion is incomplete, only generating a portion of the body instead of the full body. The second shows a teddy bear whose lower body is completed with an unreasonable and unnatural posture.}
\end{figure}

Recent works explore training-free approaches for amodal completion by utilizing powerful pre-trained generative models directly for inpainting. For instance, PD-MC\cite{xu2024amodal} employs a progressive occlusion-aware amodal completion strategy for this task. Another recent work\cite{ao2024open} utilizes large segmentation models like LISA for completing arbitrary objects specified by natural language queries, achieves impressive open-world capabilities. Nonetheless, this method\cite{ao2024open} also relies on a progressive occlusion-aware amodal completion strategy. This progressive approach, however, is inherently vulnerable to error accumulation across steps and unreliable execution, as illustrated in Fig.~\ref{fig:progressive failure examples}, compromising the consistency and fidelity of the generated results.

To overcome the limitations of the progressive strategies, we introduce a novel Collaborative Multi-Agent Reasoning Framework. Instead of sequential, step-by-step processing which suffers from error accumulation and unreliable execution, our framework orchestrates several agents. These agents collaborate in a feed-forward reasoning process to determine the complete inpainting region and semantic guidance comprehensively upfront. Our framework decouples analysis from generation, avoiding the iterative generation of progressive methods. This upfront determination of the complete inpainting region and semantic guidance, prior to the synthesis step, directly prevents the compounding of errors inherent in multi-stage generation and ensures a more reliable definition of the completion task.
Agents are dedicated to 1) segmenting visible object parts of the occluded object, 2) determining occluder-occludee relationships, and 3) determining the necessary image boundary expansion to enable the completion of truncated objects. Through the collaboration of these agents, the final region designated for inpainting is accurately defined. While this unified mask definition is efficient, it introduces a potential challenge: the resulting large inpainting area carries the risk of generating contextually related but unwanted elements, potentially leading to the regeneration of similar occluders within the designated region. To mitigate this and ensure generation fidelity to the target object, another crucial component of our framework is an agent dedicated to generating a detailed, fine-grained textual description of the occluded object, capturing properties like internal attributes and contextual information. This detailed semantic information enables Fine-Grained Semantic Guidance, guiding the inpainting model to accurately construct the desired object even within a potentially complex mask, thereby overcoming the co-occurrence issue.

A key feature of our pipeline is its direct generation of RGBA output for amodal completion in a training-free manner. Specifically, we utilize the occluded object's visible mask and attention maps from Diffusion Transformer to obtain a guidance signal, enabling the direct generation of RGBA images without requiring extra segmentation step. This design makes the completed object usable in downstream applications like augmented reality and image editing. To rigorously evaluate our solution, we performed extensive validation on multiple datasets from diverse sources. Our experimental results demonstrate that the proposed Collaborative Multi-Agent Reasoning Framework not only achieves state-of-the-art (SOTA) visual quality but also offers significant efficiency gains due to its single-pass, collaborative nature, thereby mitigating the error accumulation issues prevalent in prior progressive methods.

In summary, our main contributions are:
\begin{itemize}
\item \textbf{Collaborative Multi-Agent Reasoning Framework.} We propose a Collaborative Multi-Agent Reasoning Framework where agents collaborate within a single conceptual pass. This architecture enables complex reasoning for amodal completion while avoiding the error accumulation and inefficiency inherent in sequential progressive strategies.
\item \textbf{Multi-Agent Occlusion Analysis and Fine-Grained Description.} Within our framework, agents collaboratively perform detailed occlusion analysis and generate fine-grained, instance-specific textual descriptions. This collaborative process provides the comprehensive spatial and semantic context crucial for high-fidelity completion.
\item \textbf{Training-Free RGBA Completion.} We enable the direct generation of layered RGBA completion results in a training-free manner. This is achieved by deriving guidance signals from visible masks and attention maps, eliminating the need for subsequent segmentation steps.
\end{itemize}

\section{Related Work}
\paragraph{\textbf{Training-Based Amodal Completion}}
Amodal completion aims to infer and generate the complete structure of partially occluded objects. Existing methods broadly fall into training-based and training-free categories. Training-based approaches, including early two-stage pipelines (mask prediction then inpainting) \cite{bowen2021oconet,zhan2020self,ehsani2018segan,yan2019visualizing,zhou2021human} and later methods using generative models or self-supervision to learn priors \cite{zhan2020self,bowen2021oconet,ling2020variational}, typically require substantial task-specific data, often relying on synthetic occlusions due to the scarcity of real-world amodal annotations. Consequently, these methods generally struggle with generalization to diverse objects, handling severely occluded cases, and exhibiting robustness in out-of-distribution scenarios.
Some recent diffusion-based techniques specifically train or adapt models for amodal completion \cite{ozguroglu2024pix2gestalt,zhan2024amodal}, improving fidelity but still demanding dataset curation or model fine-tuning, inheriting limitations related to training data dependency.

\paragraph{\textbf{Training-Free Amodal Completion}}
Recent training-free approaches leverage powerful, pre-trained generative models directly, sidestepping task-specific retraining, often via multi-step inpainting procedures \cite{xu2024amodal,ao2024open}. These methods typically employ progressive strategies, iteratively generating and expanding the prediction for the occluded region. While harnessing the broad knowledge of large models enables impressive realism, this progressive nature renders them inherently vulnerable to error accumulation across steps and unreliable execution (e.g., premature termination), which compromises the consistency and fidelity of the final completion. Furthermore, defining a large inpainting region upfront, a strategy sometimes employed to avoid iterative generation, can lead the synthesis process astray; without detailed guidance, the generative model risks producing spurious content or regenerating parts of occluders within the mask, rather than the intended occluded object. Our work directly tackles the limitations of these progressive strategies through upfront, collaborative multi-agent reasoning for both spatial region definition and fine-grained semantic guidance.

\paragraph{\textbf{RGBA Image Generation}}
Early attempts at layered image generation \cite{yang2017lr,epstein2022blobgan} focused on foreground/background separation but often lacked precise alpha channel control. With the rise of diffusion models, subsequent work trained models specifically for direct RGBA image generation, typically by encoding layer information like alpha masks within the latent representation \cite{zhang2023text2layer,zhang2024transparent,hu2024diffumatting,fontanella2025generating,dalva2024layerfusion}. More recently, training-free techniques that modify the inference process, such as attention-guided alpha extraction \cite{quattrini2024alfie}, enabled the generation of transparent layers without costly model retraining. However, effectively applying these methods to amodal completion remains challenging. Accurately generating an RGBA output where the alpha channel precisely isolates only the completed object (including hallucinated parts) while excluding occluders or spurious background elements, directly from the generation process, is non-trivial and might still necessitate post-processing steps like segmentation refinement in existing pipelines. Our work contributes a method for direct, training-free RGBA generation specifically for amodal completion, utilizing guidance derived from visible masks and attention maps (from a Diffusion Transformer) to produce clean layered output without requiring subsequent segmentation.

\section{Method}
\begin{figure*}[t]
    \centering
    \includegraphics[width=0.95\textwidth]{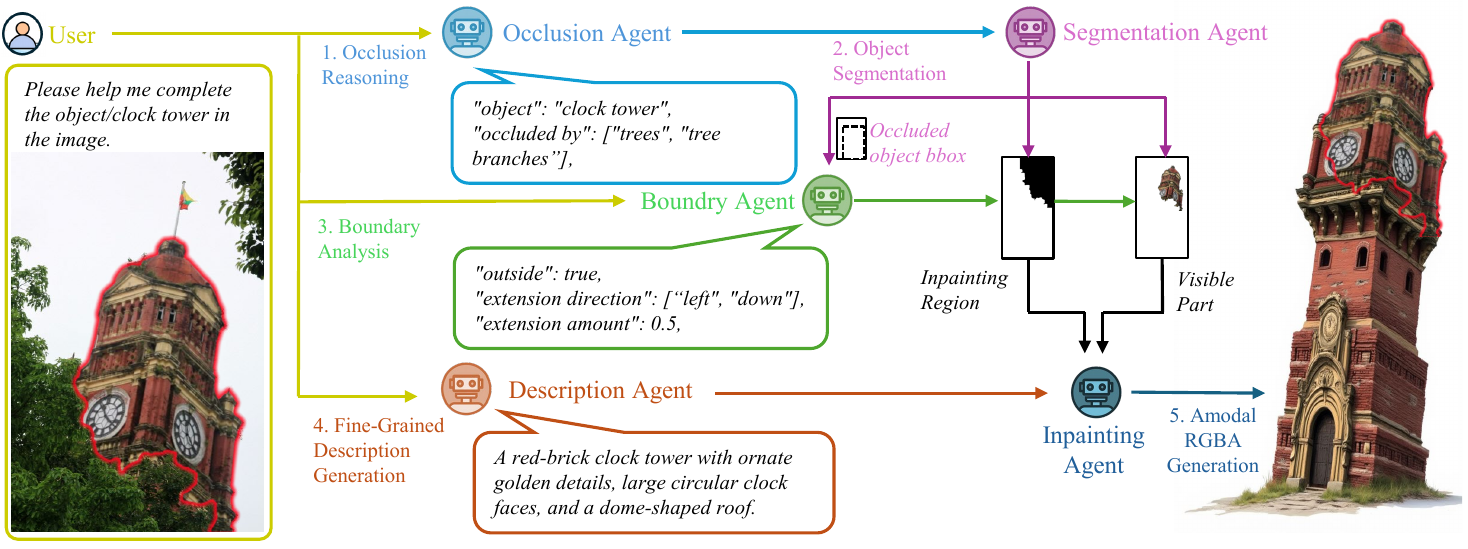}
    \caption{Overview of our Collaborative Multi-Agent Reasoning Framework. Given an input image and query, collaborating agents first perform upfront reasoning: (1) Occlusion/boundary analysis yields the inpainting mask $M_{\text{inpaint}}$. (2) A description agent generates the fine-grained prompt $P_{\text{text}}$. (3) Input $I_{\text{masked}}$ is prepared. A Inpainting Agent then uses these ($I_{\text{masked}}, M_{\text{inpaint}}, P_{\text{text}}$) to generate the completed RGB image $I_{\mathrm{complete}}$. Finally, attention maps and the visible mask $M_{\text{visible}}$ are used to generate the alpha mask $M_{\alpha}$, producing the final RGBA output.}
    \label{fig:overview}
    \Description{An overview diagram of the Collaborative Multi-Agent Reasoning Framework. The process starts with a user query, "Please help me complete the object/clock tower in the image." The query and image are sent to parallel agents. The Occlusion Agent identifies the object as "clock tower" and the occluders as "trees." The Boundary Agent determines the object is partially outside the frame and needs a 0.5 extension down and left. The Description Agent generates a detailed text: "A red-brick clock tower with ornate golden details, large circular clock faces, and a dome-shaped roof." The Segmentation Agent receives input from the Occlusion Agent, creating an inpainting mask and a visible part image. Finally, all outputs are fed into the Inpainting Agent, which performs Amodal RGBA Generation to produce the complete clock tower as a final RGBA image.}
\end{figure*}

\begin{figure}[t]
    \centering
    \includegraphics[width=0.95\columnwidth]{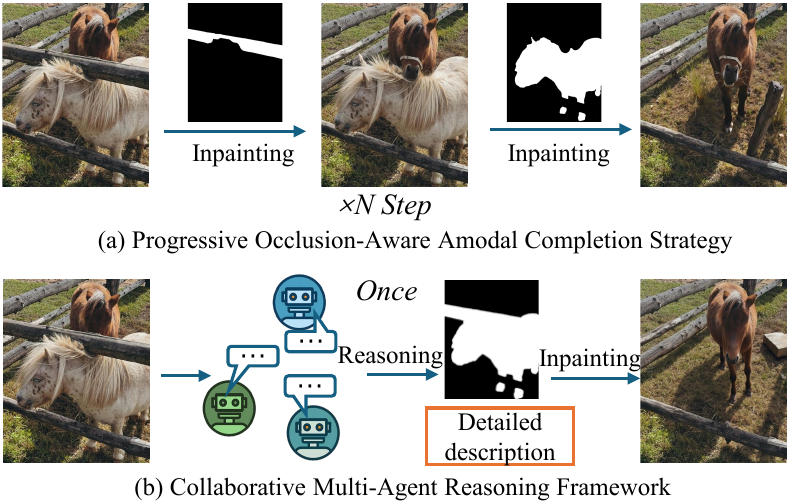}
    \caption{Comparison of amodal completion frameworks. (a) Typical progressive occlusion-aware strategies involve iterative steps for region refinement and synthesis, risking error accumulation. (b) Our Collaborative Multi-Agent Reasoning Framework employs upfront reasoning via agent collaboration to determine all spatial and semantic parameters before a single, decoupled synthesis stage, enhancing robustness.}
    \label{fig:framework_compare}
    \Description{A diagram comparing two amodal completion strategies, using an example where a white horse partially occludes a brown horse. Part (a) illustrates the "Progressive Occlusion-Aware Amodal Completion Strategy." It shows that the final result is generated through a process of continuous iteration, where the image undergoes multiple "Inpainting" steps. Part (b) illustrates our "Collaborative Multi-Agent Reasoning Framework." It shows the same input image undergoing a single "Once" process, where multiple agents engage in "Reasoning" and use a "Detailed description" to guide a single "Inpainting" stage, producing the final result in one pass.}
\end{figure}

\subsection{Framework Overview}\label{sec:overview}

We define the task of \emph{amodal completion} as follows. Given an input image $I \in \mathbb{R}^{H \times W \times 3}$, the goal is to generate an \emph{RGBA amodal completion} $I_{RGBA}$ on a potentially expanded canvas $\mathbb{R}^{H' \times W' \times 4}$. This output is an RGBA representation where the RGB channels depict the completed object (both its visible and synthesized parts), and the alpha channel ($\alpha$) isolates the object for downstream applications. The overview pipeline is shown in Fig. \ref{fig:overview}.

Achieving this requires accurately determining the occluded object's full spatial extent and semantic attributes, typically involving intermediate spatial cues (derived from visible parts $M_{\text{visible}}$, occluder masks $M_{\text{occ}}$, boundary expansion mask $\hat{M}_{\text{bdy}}$) and semantic descriptions. Prior progressive methods often suffer from error accumulation and unreliability. To address this, we propose the \textbf{Collaborative Multi-Agent Reasoning Framework}. 
Our framework decouples analysis from generation by determining the complete inpainting region and semantic guidance upfront. This single-pass reasoning avoids the iterative process of progressive methods, directly preventing the compounding of errors and ensuring a more reliable completion process
(Figure~\ref{fig:framework_compare}).

In our framework, agents first collaboratively determine the final inpainting mask $M_{\text{inpaint}} \in \{0,1\}^{H' \times W'}$ (derived from occlusion analysis, boundary expansion $\hat{M}_{\text{bdy}}$, and morphological operations, detailed in Sec.~\ref{sec:spatial_context}) and the textual semantic guidance $P_{\text{text}}$. A subsequent single synthesis pass uses a generative function $\mathcal{G}$ taking a masked input image $I_{\text{masked}}$ (containing visible object pixels on a neutral background on the potentially expanded canvas), $M_{\text{inpaint}}$, and $P_{\text{text}}$ to produce $I_{\text{complete}}$. $\mathcal{G}$ synthesizes coherent content within $M_{\text{inpaint}}$ guided by $I_{\text{masked}}$ and $P_{\text{text}}$, generating the alpha channel simultaneously. 
\begin{equation}
I_{\mathrm{complete}} = \mathcal{G}(I_{\text{masked}}, M_{\text{inpaint}}, P_{\text{text}}) \label{eq:Content_0}
\end{equation}
\begin{equation}
    I_{\mathrm{RGBA}} = \bigl[I_{\mathrm{complete}},\, \mathbf{M}_{\alpha}\bigr]. \label{eq:rgba_0}
\end{equation}

Key challenges addressed by this framework include: (1) accurately determining the object extent and semantic properties without relying on iterative refinement; (2) achieving high-fidelity synthesis harmonious with fine-grained semantic guidance while preventing contextually related but unwanted elements (like occluder regeneration) within large masks; and (3) enabling direct, training-free RGBA generation for the completed amodal object, including precise alpha masks, eliminating extra segmentation steps.

\subsection{Agent-based Spatial Reasoning}\label{sec:spatial_context}
Accurately defining the region requiring completion (\(M_{\text{inpaint}}\)) is crucial for amodal completion and challenging for prior methods relying on iterative refinement. Our Collaborative Multi-Agent Reasoning Framework addresses this through upfront, coordinated analysis performed by component agents. The reasoning process involves several coordinated steps:

\paragraph{\textbf{Occlusion Relationship Identification}}
Determining the correct occlusion relationships is fundamental to amodal completion. While prior training-free methods often use depth estimators, these can be unreliable in geometrically ambiguous cases (e.g., objects with similar depths, textureless surfaces) and prone to error propagation through iterative refinement\cite{zhan2020self}. To overcome this, our framework utilizes an occlusion identification agent endowed with advanced spatial reasoning capabilities, leveraging the power of Multimodal Large Language Models (MLLMs). We propose a novel strategy where this agent is carefully guided via specific prompting to robustly infer occlusion relationships directly from visual and contextual input. Instead of relying solely on potentially fragile geometric cues, the Occlusion agent processes the input image $I$ along with crucial context (such as an open-world query defining the target object). It is then prompted to deduce the front-back ordering between the target object and other significant scene elements. This guided reasoning process allows the agent to robustly identify the primary occluded object and enumerate its occluders ($i=1, 2, ...$), bypassing the need for iterative geometric processing.

\paragraph{\textbf{Segmentation and Boundary Analysis}} The identified occlusion relationships and object identities determined in the previous step guide the subsequent collaborative analysis by two distinct agents, which establish the detailed spatial information required for synthesis. First, a \textbf{Segmentation Agent}, employing robust open-vocabulary segmentation tools such as Grounded-Segment-Anything \cite{Ref_GroundedSAM} or LISA \cite{Ref_LISA}, generates precise pixel masks for each identified occluder, $\{M_{\text{occ}}^{(i)}\}$, and for the visible parts of the target occluded object, $M_{\text{visible}}$. 
Second, a \textbf{Boundary Analysis Agent} assesses potential object truncation by the image frame and estimates the required expansion ratio. A primary challenge here is determining truncation when the object parts near the image border are themselves occluded, rendering simple geometric checks based on visible masks insufficient. Furthermore, relying solely on direct visual inference from models for precise boundary judgments can be unreliable. Therefore, we employ a hybrid strategy. The agent first computes the bounding box (bbox) of $M_{\text{visible}}$ relative to the image dimensions $H \times W$. If the bbox touches an image edge, this provides a geometric prior signalling potential edge interaction. Crucially, to handle cases where the object might be occluded near the border and to estimate the out-of-frame extent, the agent leverages advanced reasoning capabilities of MLLMs. It takes the input image $I$ along with the geometric bbox prior (indicating edge contact) to infer the likelihood of truncation and estimate the necessary expansion. The bbox prior helps anchor the model's attention and reasoning specifically to the relevant image boundaries, enhancing the assessment of edge interaction. This combined analysis allows the agent to estimate the image boundary expansion parameters $E = \{e_l, e_r, e_t, e_b\}$, representing the relative proportions (e.g., $e_r = 0.1$ signifies expanding the width by 10\% on the right) required for each edge to likely encompass the full object.

\paragraph{\textbf{Final Inpainting Mask Generation}}
Finally, the framework combines the outputs from the agents to construct the definitive inpainting mask $M_{\text{inpaint}}$ on a potentially expanded canvas $\mathbb{R}^{H' \times W'}$. The dimensions $H', W'$ are calculated by applying the relative expansion proportions $E = \{e_l, e_r, e_t, e_b\}$ to the original dimensions $H, W$. To ensure visually coherent boundaries in the final synthesis and avoid small gaps, each occluder mask $M_{\text{occ}}^{(i)}$ first undergoes morphological expansion (dilation) on this target canvas: $\hat{M}_{\text{occ}}^{(i)} = M_{\text{occ}}^{(i)} \oplus \mathcal{B}_r$ (where $\mathcal{B}_r$ is a structuring element). A mask $\hat{M}_{\text{bdy}} \in \{0,1\}^{H' \times W'}$ representing the newly added boundary area, resulting from the application of the expansion parameters $E$, is also defined. $M_{\text{inpaint}}$ is then computed as the union of all dilated occluder masks and this boundary expansion mask:
\begin{equation}
    M_{\text{inpaint}}
    \;=\;
    \bigcup_i \hat{M}_{\text{occ}}^{(i)}
    \;\cup\;
    \hat{M}_{\text{bdy}}. \label{eq:minpaint_definition}
\end{equation}
This mask $M_{\text{inpaint}}$ precisely delineates all regions requiring synthesis. These regions include areas hidden by occluders as well as any areas extending beyond the original image frame due to the calculated boundary expansion.

\subsection{Fine-Grained Semantically Guided Synthesis}\label{sec:semantic_synthesis}

\paragraph{\textbf{Agent-based Fine-Grained Description Generation}}\label{sec:semantic_context}
Previous methods\cite{xu2024amodal} often rely on limited semantic guidance, such as simple object category labels. This coarse guidance can lead to generated content that is semantically inconsistent with the specific visible parts of the occluded object or the surrounding context. To address this crucial limitation, our framework incorporates a description agent tasked with generating a rich, fine-grained textual prompt, $P_{\text{text}}$. This agent, leveraging MLLMs, analyzes the entire input image $I$ along with context identifying the target object (e.g., an open-world query). It leverages its reasoning capabilities to infer and generate a detailed textual description hypothesizing the appearance and pose of the \emph{entire} object, including occluded portions. We employ a specific prompting strategy to guide the agent in producing this description, which encompasses visible internal attributes (like color, texture, species-specific markings), overall contextual information (like its pose), and plausible inferred characteristics of the hidden parts. For example, instead of just "cat", the agent might produce $P_{\text{text}}$ = \emph{“a full ginger tabby cat with green eyes and white paws, sitting upright and looking forward, with its tail curled around its side.”} This detailed prompt $P_{\text{text}}$, representing a reasoned hypothesis about the full object, captures far richer semantic cues than a category label, providing the necessary \textbf{Fine-Grained Semantic Guidance} for the synthesis stage.

\paragraph{\textbf{Guided Synthesis Process}}\label{sec:synthesis_process}
Before the final synthesis, we prepare the masked input image, $I_{\mathrm{masked}}$, in a process designed to isolate the target object's visible parts, minimize background interference, and accommodate necessary boundary expansion. 
First, we isolate the visible parts of the target object. Using the visible mask $M_{\mathrm{visible}}$ (determined in Sec.~\ref{sec:spatial_context}), we create an intermediate image, $I_{\mathrm{vis\_only}} \in \mathbb{R}^{H \times W \times 3}$, where pixels corresponding to $M_{\mathrm{visible}}$ retain their original RGB values from $I$, while all other pixels are set to a neutral background color (e.g., white).
Second, we place the isolated object onto the potentially expanded canvas. Using the boundary expansion parameters $E$ (also determined in Sec.~\ref{sec:spatial_context}), we determine the final canvas dimensions $H' \times W'$. We then construct $I_{\mathrm{masked}} \in \mathbb{R}^{H' \times W' \times 3}$ by placing the content of $I_{\mathrm{vis\_only}}$ at its original position within this new canvas and filling any newly added areas with the neutral background color.

The final synthesis is performed by the Inpainting Agent, which executes the generative function $\mathcal{G}$ using three key inputs from the prior agent collaboration: 
1) The prepared masked input image $I_{\mathrm{masked}} \in \mathbb{R}^{H' \times W' \times 3}$.
2) The comprehensive inpainting mask $M_{\mathrm{inpaint}} \in \{0,1\}^{H' \times W'}$. 
3) The detailed textual prompt $P_{\mathrm{text}}$. 
Guided by the fine-grained semantics in $P_{\mathrm{text}}$ and conditioned on the limited visual context of the object itself provided by $I_{\mathrm{masked}}$, the Inpainting Agent generates the completed RGB image $I_{\mathrm{complete}} \in \mathbb{R}^{H' \times W' \times 3}$ within the specified $M_{\text{inpaint}}$ region:
\begin{equation}
I_{\mathrm{complete}} = \mathcal{G}\Big(I_{\mathrm{masked}},\, M_{\mathrm{inpaint}},\, P_{\mathrm{text}}\Big). \label{eq:synthesis}
\end{equation}
This Fine-Grained Semantic Guidance ensures the synthesized content aligns accurately with the specific characteristics of the occluded object and avoids the generation of contextually related but unwanted elements.

\subsection{Training-Free Amodal RGBA Generation}\label{sec:rgba_generation}

A key goal of our framework is to directly generate a complete RGBA representation $I_{\mathrm{RGBA}} = [I_{\mathrm{complete}}, \mathbf{M}_{\alpha}]$ of the amodally completed object, including a precise alpha mask $\mathbf{M}_{\alpha}$, without requiring separate post-processing segmentation. Recent training-free methods have shown promise in estimating alpha masks by aggregating attention maps from generative models like Diffusion Transformers \cite{quattrini2024alfie}. However, directly applying such techniques to \emph{amodal completion} presents a unique challenge. Cross-attention maps, which link generated pixels to text prompts, often strongly highlight the newly synthesized (previously occluded) regions but may provide weaker or incomplete signal over the object's original visible parts present in the input $I_{\text{masked}}$. Relying solely on attention aggregation might therefore yield an incomplete alpha mask covering only the hallucinated portions.

To overcome this limitation and obtain a complete mask for the entire object (visible plus synthesized parts), our core strategy is to fuse the attention-derived mask with the known visible mask $M_{\text{visible}}$ (obtained during initial analysis, Sec.~\ref{sec:spatial_context}). This fusion ensures that both the accurately known visible geometry and the newly synthesized parts guided by attention are captured in the final alpha channel.

\paragraph{\textbf{Attention-Based Object Mask}}
The Inpainting Agent, using the Flux model, utilizes Multi-stream blocks employing cross-attention (image $\leftrightarrow$ text) and self-attention (image $\leftrightarrow$ image). Let $\mathbf{A}_{t,l}^{C}$ be the cross-attention map at time $t$, layer $l$. Following prior work \cite{cao2023masactrl,chefer2023attend,tewel2024training}, we identify latent features corresponding to the target object by aggregating cross-attention maps related to the subject tokens in the detailed prompt $P_{\text{text}}$. Considering that later denoising steps capture finer spatial details, we typically average maps from the final 15 steps across relevant layers to obtain a global cross-attention map $\mathbf{A}^{C}$. Thresholding $\mathbf{A}^{C}$ yields a coarse object mask $\mathbf{M}_{\mathrm{C}}$, primarily highlighting the synthesized regions. As motivated above, we fuse this attention-based mask with the known visible mask $M_{\text{visible}}$ to get a complete initial estimate: $\mathbf{M}_{\mathrm{fused}} = \mathbf{M}_{\mathrm{C}} \cup M_{\text{visible}}$.

\paragraph{\textbf{Refinement and Final RGBA Composition}}
Self-attention maps $\mathbf{A}_{t,l}^{S}$ are also aggregated to help refine the boundaries of the fused mask ($\mathbf{M}_{\mathrm{fused}}$). A lightweight refinement step (e.g., GrabCut \cite{Ref_GrabCut}), initialized with this refined estimate, produces the final alpha mask $\mathbf{M}_{\alpha} \in \{0,1\}^{H' \times W'}$. This mask is then combined with the synthesized RGB image $I_{\mathrm{complete}}$ (Eq.~\ref{eq:synthesis}) to yield the final layered RGBA output:
\begin{equation}
    I_{\mathrm{RGBA}} = \bigl[I_{\mathrm{complete}},\, \mathbf{M}_{\alpha}\bigr]. \label{eq:rgba}
\end{equation}
This process delivers a clean object representation suitable for downstream tasks without further segmentation.
\section{Experiments}\label{sec:experiments}
\subsection{Datasets, Metrics and Implementation Details}
\paragraph{\textbf{Datasets}}
Evaluating amodal completion is challenging due to the lack of ground truth and dataset limitations. For evaluation, we utilize two main data sources. First, for open-world scenarios and comparison, we employ the benchmark methodology proposed by Ao et al. \cite{ao2024open}, excluding its LAION dataset component due to data access instability. Second, to increase data diversity and cover more varied real-world conditions, we additionally curated a custom dataset comprising everyday photographs taken by us and royalty-free images collected from the internet. 

\paragraph{\textbf{Metrics}}
Due to the inherent unavailability of ground-truth appearances of occluded regions, we rely on a combination of subjective human evaluation and quantitative metrics computed on visible regions.
For subjective assessment, human evaluators rate completions based on Object Completeness, Visual Coherence, and Overall Plausibility. For quantitative analysis, following Ao et al.~\cite{ao2024open}, we report metrics measuring semantic alignment and visual consistency by comparing the \emph{visible} parts of the original object with the corresponding regions in the completed image $I_{\mathrm{complete}}$. These include \textbf{CLIP score}~\cite{radford2021learning} (measuring semantic alignment between the completed visible parts and the object category), \textbf{LPIPS}~\cite{zhang2018unreasonable} (perceptual similarity), \textbf{VGG16 feature similarity}~\cite{gatys2016image} (deep feature consistency), and \textbf{SSIM}~\cite{wang2004image} (structural similarity).

\paragraph{\textbf{Implementation Details}}
Our Framework operates in a training-free manner, requiring no task-specific model training or fine-tuning. For the agents performing multimodal reasoning, we utilize OpenAI’s GPT-4o \cite{Ref_GPT4o} as the core engine.  For the Segmentation Agent, we employ Grounded-Segment-Anything \cite{Ref_GroundedSAM}. For the Inpainting Agent, we adopt FLUX-ControlNet-Inpainting model \cite{Ref_FluxControlNetInpainting}. All experiments were conducted on NVIDIA A800 GPUs. When comparing with prior approaches, we utilized their official public codebases and pre-trained models as released by the authors, following their recommended evaluation protocols.

\begin{table}[t] 
\caption{Quantitative comparison with SOTA methods on the Ao et al.~\cite{ao2024open} benchmark. Baselines are from \cite{ao2024open} and ours are evaluated under identical conditions. All metrics assess visible part consistency. $\uparrow$: higher is better, $\downarrow$: lower is better.}
\label{tab:results_comparison_updated}
  \centering
  \resizebox{\columnwidth}{!}{%
    \begin{tabular}{@{}llcccc@{}} 
      \toprule
      \multirow{2}{*}{Dataset} & \multirow{2}{*}{Method} & Class & Visual & Semantic & Structural \\
      & & Relevance & Consistency & Consistency & Consistency \\
      \cmidrule(lr){3-3} \cmidrule(lr){4-4} \cmidrule(lr){5-5} \cmidrule(lr){6-6}
      & & ↑ CLIP & ↓ LPIPS & ↑ Feature Sim. & ↑ SSIM \\
      \midrule
      \multirow{5}{*}{VG} 
        & PD w/o MC\cite{xu2024amodal}   & 28.254 & 0.586 & 0.406 & 0.459 \\
        & PD-MC\cite{xu2024amodal}       & 28.367 & 0.578 & 0.413 & 0.463 \\
        & Pix2gestalt\cite{ozguroglu2024pix2gestalt} & 27.672 & 0.429 & 0.554 & 0.726 \\
        & OWAAC\cite{ao2024open}       & \textbf{28.470} & 0.310 & 0.658 & 0.732 \\ 
        \rowcolor{highlightgray} 
        & Ours        & 28.011  & \textbf{0.218} & \textbf{0.855} & \textbf{0.839} \\ 
      \midrule
      \multirow{5}{*}{COCO-A}
        & PD w/o MC\cite{xu2024amodal}   & 27.426 & 0.673 & 0.318 & 0.381 \\
        & PD-MC\cite{xu2024amodal}       & 27.383 & 0.664 & 0.328 & 0.382 \\
        & Pix2gestalt\cite{ozguroglu2024pix2gestalt} & 26.998 & 0.471 & 0.524 & 0.695 \\
        & OWAAC\cite{ao2024open}       & \textbf{27.612} & 0.351 & 0.609 & 0.718 \\ 
        \rowcolor{highlightgray}
        & Ours        & 27.572  & \textbf{0.250} & \textbf{0.831} & \textbf{0.820} \\ 
      \midrule
      \multirow{5}{*}{Free Images} 
        & PD w/o MC\cite{xu2024amodal}   & 28.190 & 0.730 & 0.268 & 0.305 \\
        & PD-MC\cite{xu2024amodal}       & 28.333 & 0.720 & 0.279 & 0.309 \\
        & Pix2gestalt\cite{ozguroglu2024pix2gestalt} & 27.621 & 0.393 & 0.613 & 0.732 \\
        & OWAAC\cite{ao2024open}       & \textbf{28.652} & 0.269 & 0.698 & 0.753 \\ 
        \rowcolor{highlightgray} 
        & Ours        & 28.316 & \textbf{0.218} & \textbf{0.847} & \textbf{0.840}  \\ 
      \bottomrule
    \end{tabular}
  }
\end{table}

\begin{figure*}[t]
    \centering
    \includegraphics[width=0.85\textwidth]{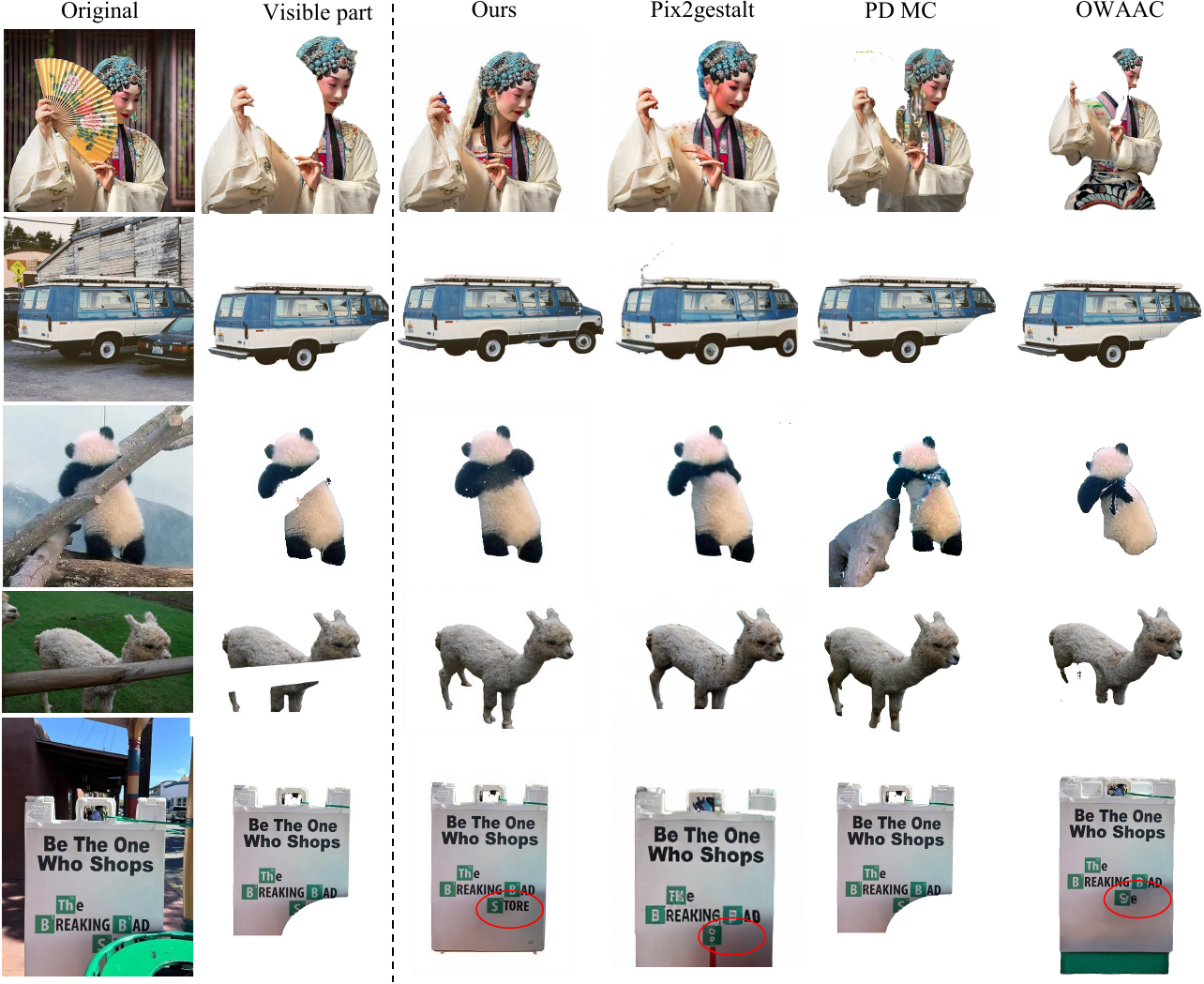}
    \caption{Qualitative comparison of our method against SOTA approaches Pix2Gestalt\cite{ozguroglu2024pix2gestalt}, PD-MC\cite{xu2024amodal}, and OWAAC\cite{ao2024open}. Note that methods employing progressive occlusion-aware strategies, such as PD-MC and OWAAC, sometimes exhibit significant artifacts or outright completion failures in challenging cases, likely due to error propagation. Our Collaborative Multi-Agent Reasoning Framework consistently produces more robust, coherent, and visually plausible results across diverse scenarios.}
    \label{fig:main_result}
    \Description{A qualitative comparison of different amodal completion methods across five challenging examples, shown in rows. The columns display the "Original" image, the "Visible part" given to the models, and the results from "Ours," "Pix2gestalt," "PD MC," and "OWAAC." For instance, in the third row, the original image shows a panda partially occluded by a fence. The results from PD MC and OWAAC show visible artifacts or incomplete shapes. In contrast, the result from the "Ours" column consistently shows a more complete, coherent, and artifact-free panda. This pattern of higher quality and robustness for our method is demonstrated across other examples, including a person in an opera costume, a van, an alpaca, and a sign with text.}
\end{figure*}

\subsection{Comparisons with Other Methods}

We conducted comprehensive comparisons of our Collaborative Multi-Agent Reasoning Framework against several SOTA amodal completion approaches, including Pix2Gestalt \cite{ozguroglu2024pix2gestalt}, PD-MC \cite{xu2024amodal}, and OWAAC \cite{ao2024open}. Qualitative and quantitative results highlight the advantages of our method.
As illustrated in Fig.~\ref{fig:main_result}, notable performance differences emerge. Methods relying on progressive, iterative strategies, such as PD-MC \cite{xu2024amodal} and OWAAC \cite{ao2024open}, exhibit limitations. For instance, PD-MC can struggle with correctly handling complex occlusions or completing the inpainting task successfully in some scenarios. We attribute such issues, inherent in multi-step refinement approaches, primarily to the risk of error accumulation and unreliable execution during the iterative process. Separately, Pix2Gestalt \cite{ozguroglu2024pix2gestalt} performs reasonably for simpler objects but often shows reduced effectiveness when dealing with images containing more complex content or intricate object details.
In contrast, our approach consistently yields superior results across diverse test cases. By employing upfront collaborative reasoning among agents to determine spatial context and leveraging Fine-Grained Semantic Guidance for the synthesis stage, our method robustly handles complex occlusions, accurately completes hidden regions, and effectively manages object truncation via boundary expansion, even for visually intricate scenes. 
This leads to more coherent and faithful completions, demonstrating the enhanced robustness and accuracy of our non-iterative pipeline over prior progressive strategies.

\begin{table}[ht] 
\caption{Quantitative comparison with SOTA methods on the Internet and Daily-Life datasets. Metrics assess consistency between original visible parts and completed regions.}
\label{tab:main_results}
    \centering
    \resizebox{\columnwidth}{!}{%
    \begin{tabular}{llcccc}
      \toprule
      \multirow{2}{*}{Dataset} & \multirow{2}{*}{Method} & Class & Visual & Semantic & Structural \\
      & & Relevance & Consistency & Consistency & Consistency \\
      \cmidrule(lr){3-3} \cmidrule(lr){4-4} \cmidrule(lr){5-5} \cmidrule(lr){6-6}
      & & ↑ CLIP & ↓ LPIPS & ↑ Feature Sim. & ↑ SSIM \\
      \midrule
        \multirow{4}{*}{Internet} 
            & Pix2Gestalt \cite{ozguroglu2024pix2gestalt} & 27.122 & 0.207 & 0.774 & 0.880 \\
            & PD-MC \cite{xu2024amodal}         & 25.682 & 0.451 & 0.675 & 0.730 \\
            & OWAAC \cite{ao2024open}           & 27.187 & 0.408 & 0.743 & 0.748 \\ 
            \rowcolor{highlightgray}
            & \textbf{Ours}                     & \textbf{27.303} & \textbf{0.151} & \textbf{0.853} & \textbf{0.910} \\
        \midrule
        \multirow{4}{*}{Daily-Life} 
            & Pix2Gestalt \cite{ozguroglu2024pix2gestalt} & 26.892 & 0.147 & 0.792 & 0.920 \\
            & PD-MC \cite{xu2024amodal}         & 26.150 & 0.426 & 0.633 & 0.770 \\
            & OWAAC \cite{ao2024open}           & \textbf{27.836} & 0.416 & 0.784 & 0.725 \\ 
            \rowcolor{highlightgray} 
            & \textbf{Ours}                     & 27.384 & \textbf{0.094} & \textbf{0.916} & \textbf{0.943} \\
        \bottomrule
    \end{tabular}
    }
\end{table}

In terms of efficiency, our framework (25.51s on an A800 GPU) is substantially faster than progressive methods like PD-MC\cite{xu2024amodal} (43.30s) and OWAAC\cite{ao2024open} (51.82s). 
While Pix2Gestalt\cite{ozguroglu2024pix2gestalt} (17.53s) is faster, our method's training-free flexibility avoids costly retraining and allows for integrating models like Qwen2.5-VL\cite{bai2025qwen2} to achieve a competitive 24.15s runtime.

\subsection{Ablation Studies}\label{sec:ablation}

\begin{figure}[t]
    \centering
    \includegraphics[width=1\columnwidth]{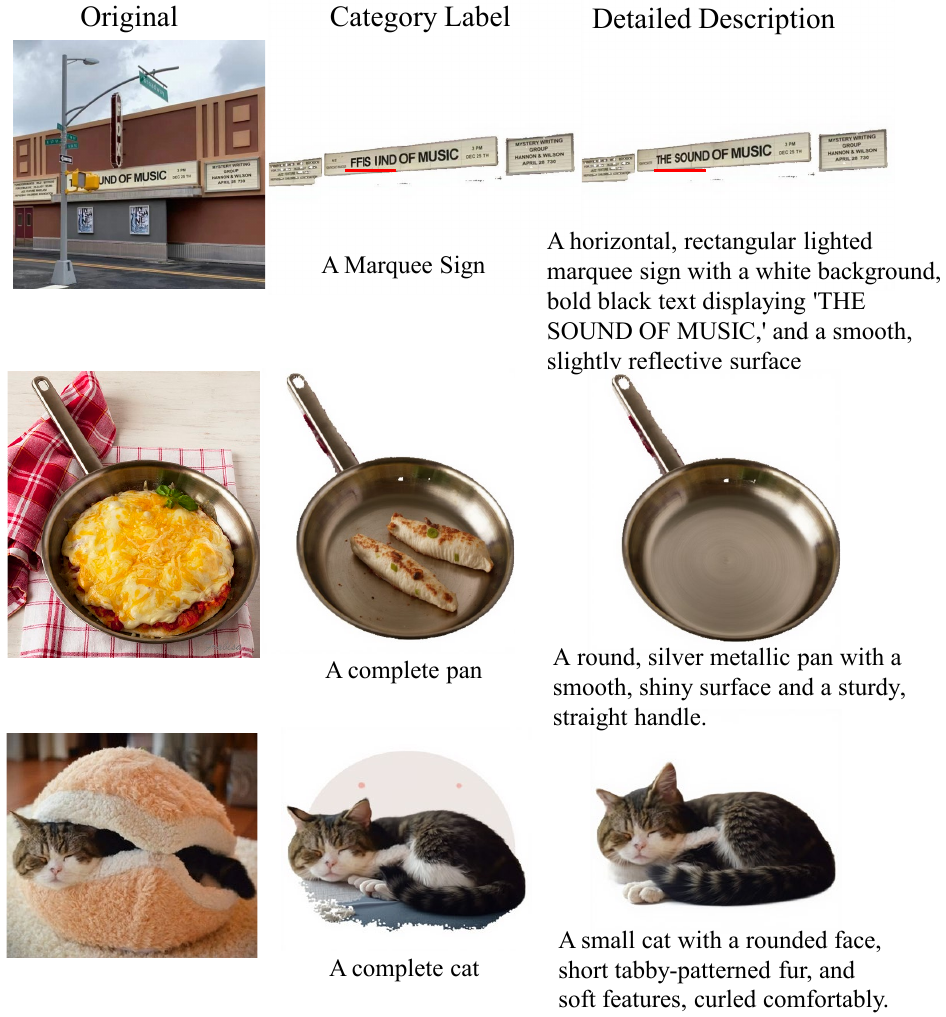}
    \caption{Fine-Grained Semantic Guidance enables superior amodal completion. Unlike simple category labels, our detailed descriptions accurately restore specifics (e.g., occluded text) and prevent artifacts.}
    \label{fig:ablation_guidance}
    \Description{An ablation study comparing amodal completion using a simple "Category Label" versus a "Detailed Description." The figure has three rows, each showing an original image and the results from these two guidance methods. In the first row, the original image is a marquee sign with occluded text. The "Category Label" method generates a generic sign, while the "Detailed Description" method accurately restores the text "THE SOUND OF MUSIC." In the second row, for an occluded pan, the "Category Label" method generates a pan containing unexpected objects, such as fried dumplings, instead of an empty pan. In contrast, the "Detailed Description" method correctly generates a clean, empty silver pan. In the third row, for a partially hidden cat, the detailed description produces a more accurate and contextually appropriate completion than the generic category label.}
\end{figure}

\begin{figure}[t]
    \centering
    \includegraphics[width=0.95\columnwidth]{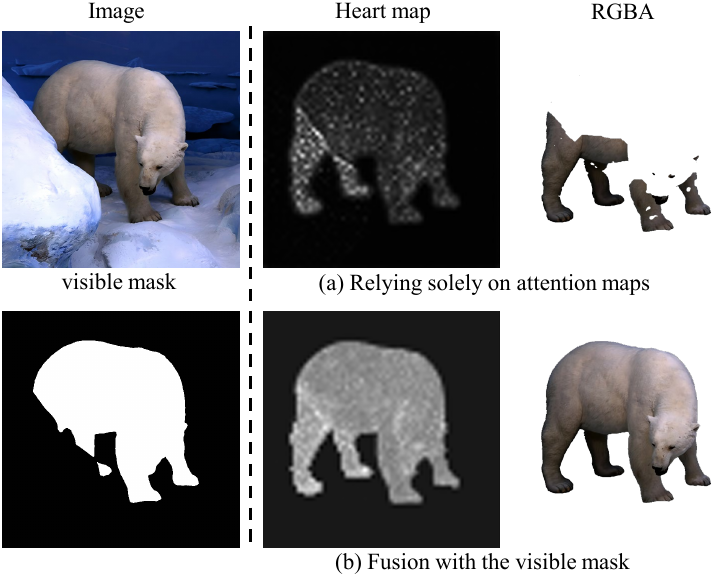}
    \caption{Ablation for RGBA mask generation. (a) Using only attention maps produces an incomplete mask, failing on visible regions. (b) Our fusion with the visible mask yields a complete and accurate alpha channel.}
    \label{fig:ablation_rgba}
    \Description{An ablation study for RGBA mask generation, comparing two approaches. Row (a), titled "Relying solely on attention maps," shows that using only the heat map from the model results in an incomplete alpha mask for a polar bear; the mask only covers the newly synthesized parts and fails to include the original visible regions, resulting in a fragmented object. Row (b), titled "Fusion with the visible mask," demonstrates our proposed method. By fusing the attention-based heat map with the known visible mask of the polar bear, the final generated RGBA output contains a complete and accurate alpha channel that correctly outlines the entire object, including both the original and the completed parts.}
\end{figure}

\begin{table}[ht] 
\caption{Quantitative ablation study results.}
    \label{tab:ablation}
    \centering
    \resizebox{\columnwidth}{!}{
    \begin{tabular}{@{}lcccc@{}}
        \toprule
        \multirow{2}{*}{Method Variant} & Class & Visual & Semantic & Structural \\
        & Relevance & Consistency & Consistency & Consistency \\
        \cmidrule(lr){2-2} \cmidrule(lr){3-3} \cmidrule(lr){4-4} \cmidrule(lr){5-5}
        & ↑ CLIP & ↓ LPIPS & ↑ Feature Sim. & ↑ SSIM \\
        \midrule
        Baseline (Single Agent)        & 27.130           & 0.221           & 0.836            & 0.845            \\
        w/o Detailed Description         & 27.312           & 0.172           & 0.843            & 0.879            \\
        w/o Boundary Analysis        & \textbf{27.641}           & 0.175           & 0.834           & 0.897            \\
        w/o Occlusion Reasoning         & 27.178           & 0.329           & 0.688            & 0.795            \\
        GPT-4o\cite{Ref_GPT4o} -> Qwen2.5-VL\cite{bai2025qwen2} &27.198 & 0.194& 0.827& 0.861 \\
        \rowcolor{highlightgray}
        Full Framework          & 27.351 & \textbf{0.127}  & \textbf{0.883}   & \textbf{0.925}   \\
        \bottomrule
    \end{tabular}
    } 
\end{table}

We perform ablation studies to validate the core components of our Collaborative Multi-Agent Reasoning Framework. 
We evaluate the impact of core components like multi-agent collaboration and fine-grained semantic guidance, and also demonstrate the framework's flexibility by testing alternative components, such as a different reasoning agent (Table~\ref{tab:ablation}, Fig.~\ref{fig:ablation_guidance}).
The role of visible mask fusion for RGBA generation is shown qualitatively in Fig.~\ref{fig:ablation_rgba}. 
Comparing our full framework against a single-agent baseline confirms that decomposing complex reasoning among collaborating agents yields significantly more robust and accurate parameter determination (Table~\ref{tab:ablation}). Furthermore, replacing detailed prompts ($P_{\text{text}}$) with category labels validates that Fine-Grained Semantic Guidance is essential for semantic consistency and accurately synthesizing specific details (e.g., text on posters), preventing unwanted element generation within large masks (Fig.~\ref{fig:ablation_guidance}). Lastly, qualitative results demonstrate that generating the RGBA alpha mask ($M_{\alpha}$) solely from attention maps leads to incomplete coverage of visible parts; fusing these maps with the known visible mask ($M_{\text{visible}}$) is necessary for a complete and precise alpha channel (Fig.~\ref{fig:ablation_rgba}). These studies collectively confirm the significant contribution of the evaluated components to the framework's effectiveness.

\subsection{User Study}\label{sec:user_study}
To evaluate subjective perceptual quality, which automatic metrics cannot fully capture, we conducted a user study with 30 participants comparing our method against Pix2Gestalt \cite{ozguroglu2024pix2gestalt}, PD-MC \cite{xu2024amodal}, and OWAAC \cite{ao2024open}. Participants viewed the original image followed by the randomized completion results from the four methods. They rated each result independently on a 5-point Likert scale (1=Very Poor, 5=Very Good) based on three criteria: (1) Object Completeness (How well does the generated region structurally complete the object's physical form where hidden?), (2) Visual Coherence (How seamlessly do generated parts merge with visible parts?), and (3) Overall Plausibility (How realistic and believable does the entire completed object appear within the scene context?). As summarized in Table~\ref{tab:user_study}, our method consistently received significantly higher average ratings across all criteria, demonstrating a clear user preference and confirming the perceptual benefits derived from our framework's robustness and fine-grained guidance.

\subsection{Limitations}
Our framework's performance is limited by the segmentation agent, as inaccuracies can cause visual artifacts or instance ambiguity. However, the framework’s modularity provides a clear path to overcome this limitation, as more advanced segmentation agents can be seamlessly integrated in the future.

\begin{table}[t] 
\caption{User study results comparing our method against baselines. Values are average scores on a 1-5 Likert scale (higher is better), based on 30 participants.}
    \label{tab:user_study}
    \centering
    
    \resizebox{\columnwidth}{!}{%
    \begin{tabular}{lcccc}
        \toprule
        Method                             & Completeness ($\uparrow$) & Coherence ($\uparrow$) & Plausibility ($\uparrow$) & Average ($\uparrow$) \\
        \midrule
        Pix2Gestalt \cite{ozguroglu2024pix2gestalt} & 3.27                    & 3.37                   & 3.40                      & 3.35                 \\
        PD-MC \cite{xu2024amodal}          & 2.83                    & 3.34                   & 2.93                      & 3.03                 \\
        OWAAC \cite{ao2024open}             & 3.20                    & 3.50                   & 3.43                      & 3.38                 \\
        \textbf{Ours}     & \textbf{3.53}           & \textbf{3.87}          & \textbf{3.60}             & \textbf{3.67}        \\
        \bottomrule
    \end{tabular}
    }
\end{table}

\section{Conclusion}
This work presents a novel Collaborative Multi-Agent Reasoning Framework that overcomes the error accumulation and unreliability of prior progressive methods in amodal completion. By decoupling analysis from synthesis via an upfront, multi-agent reasoning process, our non-iterative approach significantly enhances robustness. The framework leverages fine-grained semantic guidance for high-fidelity synthesis and directly generates layered RGBA outputs, eliminating the need for subsequent segmentation. Extensive evaluations confirm our method achieves state-of-the-art visual quality, representing a practical step towards generating high-quality digital assets for creative applications.



\begin{acks}
This work was supported by \grantsponsor{nsfc}{National Natural Science Foundation of China}{} (\grantnum{nsfc}{62132001}), \grantsponsor{cfh}{Capital's Funds for Health Improvement and Research}{} (\grantnum{cfh}{CFH 2024-2-40611}), and the \grantsponsor{frfcu}{Fundamental Research Funds for the Central Universities}{}.
\end{acks}

\bibliographystyle{ACM-Reference-Format}
\bibliography{main}

\clearpage
\appendix
\appendix
\section{Agent Implementation and Prompting Strategies}\label{sec:appendix_agents_prompts}

\begin{figure}[t]
    \centering
    \includegraphics[width=0.9\columnwidth]{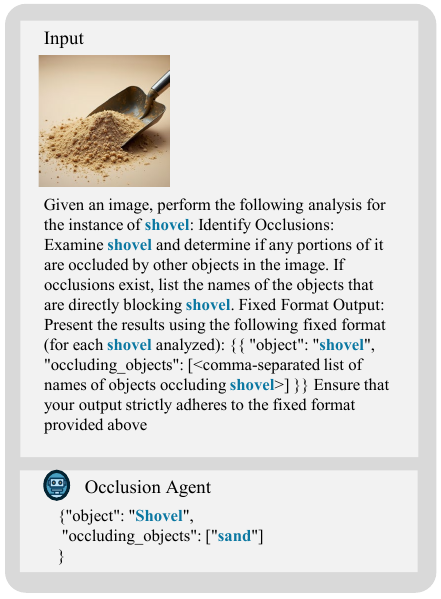}
    \caption{Example interaction with the Occlusion Identification Agent.}
    \label{fig:id_agent_prompt_example}
\end{figure}

\begin{figure}[t]
    \centering
    \includegraphics[width=0.9\columnwidth]{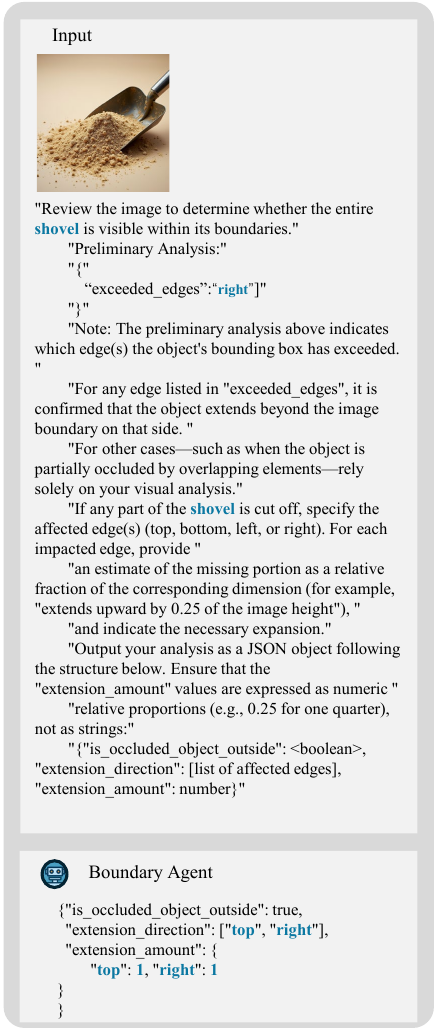}
    \caption{Example interaction with the Boundary Analysis Agent.}
    \label{fig:boundary_agent_prompt_example}
\end{figure}

\begin{figure}[t]
    \centering
    \includegraphics[width=0.9\columnwidth]{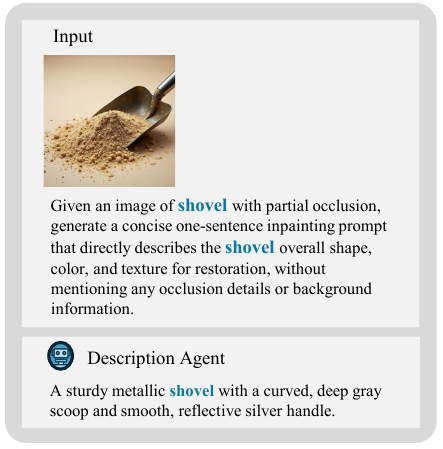}
    \caption{Example interaction with the Description Agent.}
    \label{fig:desc_agent_prompt_example}
\end{figure}

This section provides further implementation details for the agents within our Collaborative Multi-Agent Reasoning Framework. Specific prompt structures and input/output examples for the reasoning agents are illustrated in Figure~\ref{fig:id_agent_prompt_example} (Identification), Figure~\ref{fig:boundary_agent_prompt_example} (Boundary Analysis), and Figure~\ref{fig:desc_agent_prompt_example} (Description).

\paragraph{Identification Agent}
Our reasoning agents leverage the capabilities of powerful Multimodal Large Language Models (MLLMs), such as OpenAI's GPT-4o \cite{Ref_GPT4o}. The Identification Agent takes the input image $I$ and a textual query specifying the target object (this supports specific queries like "the cat sleeping on the chair" or more open-ended descriptions like "the red object near the center"). It utilizes a structured prompting strategy (details in Fig.~\ref{fig:id_agent_prompt_example}) to guide the MLLM in identifying the primary occluded object and enumerating its occluders based on spatial relationship inference. Its output is a parsed list of occluder identities.

\paragraph{Segmentation Agent}
The Segmentation Agent uses the input image $I$ and the identities provided by the Identification Agent. Employing standard usage of robust open-vocabulary segmentation tools like Grounded-Segment-Anything \cite{Ref_GroundedSAM} or LISA \cite{Ref_LISA}, it generates the pixel masks $M_{\text{visible}}$ (for the target's visible parts) and $\{M_{\text{occ}}^{(i)}\}$ (for each occluder).

\paragraph{Boundary Analysis Agent}
This agent, also leveraging an MLLM (typically GPT-4o), assesses potential object truncation. Taking the image $I$ and visible mask $M_{\text{visible}}$ as input, it first computes a geometric prior based on bounding box contact with image edges. This prior, along with the image, is used to prompt the MLLM (see Fig.~\ref{fig:boundary_agent_prompt_example}) for a reasoned estimation of truncation likelihood and the proportional expansion parameters $E = \{e_l, e_r, e_t, e_b\}$ needed for each edge.

\paragraph{Description Agent}
The Description Agent utilizes an MLLM (such as GPT-4o) to generate the fine-grained textual guidance $P_{\text{text}}$. Taking the image $I$ and target object context (e.g., the user-provided query) as input, it follows a prompting strategy (examples in Fig.~\ref{fig:desc_agent_prompt_example}) that guides the MLLM to first describe visible attributes and pose, then infer and include plausible characteristics of hidden parts, producing a description of the hypothesized complete object.

\paragraph{Synthesis Agent}
For the Synthesis Agent, we employ the FLUX-ControlNet-Inpainting model \cite{Ref_FluxControlNetInpainting}. This agent receives the prepared masked image $I_{\text{masked}}$, the comprehensive inpainting mask $M_{\text{inpaint}}$, and the detailed prompt $P_{\text{text}}$ as input. Key inference parameters include 28 denoising steps. Its output is the completed RGB image $I_{\mathrm{complete}}$.

\paragraph{RGBA Generation Details}
Finally, the direct RGBA generation process aggregates cross-attention ($\mathbf{A}^{C}$) and self-attention ($\mathbf{A}^{S}$) maps from the final 10 steps of the Synthesis Agent (Flux). The thresholded $\mathbf{A}^{C}$ ($\mathbf{M}_{\mathrm{C}}$) is fused with the visible mask $M_{\text{visible}}$ ($\mathbf{M}_{\mathrm{fused}} = \mathbf{M}_{\mathrm{C}} \cup M_{\text{visible}}$). This fused mask is refined using $\mathbf{A}^{S}$, followed by GrabCut \cite{Ref_GrabCut} to produce the final alpha mask $\mathbf{M}_{\alpha}$. The final output $I_{\mathrm{RGBA}}$ is composed as $[I_{\mathrm{complete}}, \mathbf{M}_{\alpha}]$.

\section{Additional Qualitative Results}
\begin{figure*}[t]
    \centering
    \includegraphics[width=0.9\textwidth]{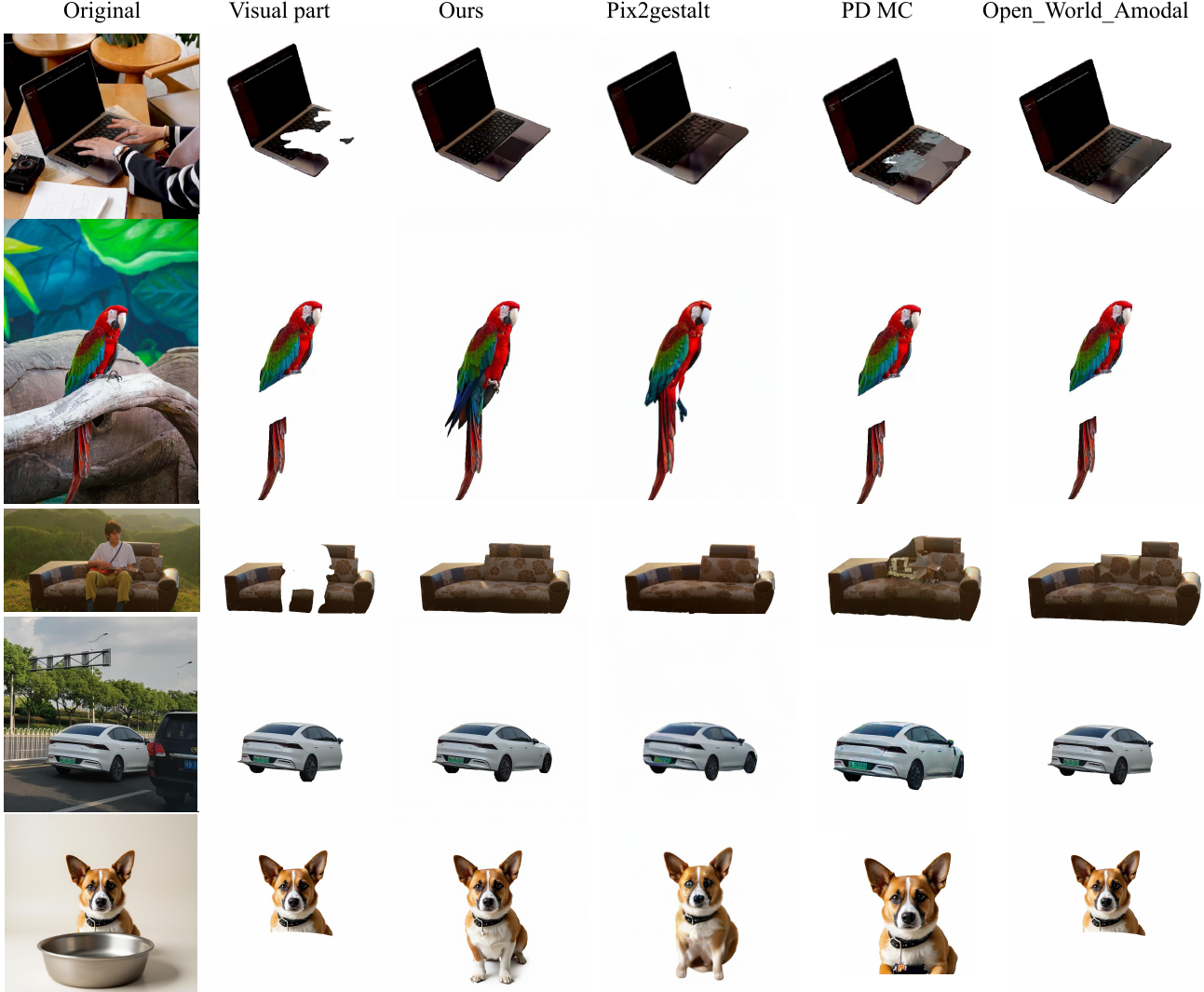}
    \caption{Additional qualitative results of our method on diverse scenarios.}
    \label{fig:more_result}
\end{figure*}

This section provides supplementary qualitative results to further demonstrate the capabilities and robustness of our proposed Collaborative Multi-Agent Reasoning Framework. Figures~\ref{fig:more_result} showcase additional examples of amodal completion across a diverse range of object categories, challenging real-world occlusion configurations, and scenarios involving significant boundary truncation. These results complement those presented in the main paper (Sec.~\ref{sec:experiments}) and provide further visual evidence of our method's ability to generate high-fidelity, semantically coherent, and plausible completions, particularly in comparison to baseline approaches on difficult cases.

\section{More Ablation Studies and Analysis}
\subsection{Ablation on Occlusion Reasoning Approach}
\label{sec:appendix_occlusion_ablation} 

We investigate the impact of our collaborative multi-agent structure specifically on the crucial task of occlusion relationship prediction. Table~\ref{tab:occlusion_reasoning_ablation} compares the performance of a baseline using a single agent (prompting GPT-4o directly) against our full collaborative framework employing different powerful Multimodal Large Language Models (MLLMs) as the reasoning engine(s). Performance is measured by overall prediction Accuracy and Hallucination Rate (predicting occluders when none exist).

The results clearly demonstrate the significant advantage of the collaborative framework. Compared to the 'Single Agent (GPT-4o only)' baseline, our framework using the same GPT-4o model ('Ours (Agents w/ GPT-4o)') achieves a notable increase in Accuracy (92.5\% vs 87.2\%) and more than halves the Hallucination Rate (5.1\% vs 10.5\%). This suggests that decomposing the reasoning and potentially allowing for internal consistency checks or more structured prompting within the multi-agent setup leads to substantially more reliable occlusion identification than directly prompting a single model for the complex task.

We also observe performance variations based on the underlying MLLM within our framework. While Gemini 1.5 Pro also performs well (91.5\% Accuracy, 8.2\% Hallucination), significantly outperforming the single-agent baseline, GPT-4o yielded the best results in our experiments for this specific task. This highlights that while the collaborative architecture provides major benefits, the choice of the foundational MLLM can further influence performance. Nevertheless, the primary finding is the consistent superiority of the multi-agent collaborative approach over the single-agent baseline.

\subsection{Ablation on Boundary Analysis Strategy}

Table~\ref{tab:ablation_boundary_detail} presents an ablation study evaluating different strategies employed by the Boundary Analysis Agent for predicting the necessity and suitability of image boundary expansion when objects might be truncated. We compare three approaches: a simple heuristic based only on whether the visible object's bounding box touches the image edge ('Baseline: Bbox Touching Edge Only'), agent-based reasoning using only the input image ('Agent Reasoning (Image Only)'), and our proposed hybrid method using both the image and the bounding box prior ('Ours (Image + Bbox Prior)'). Accuracy in determining the correct edge(s) for expansion ('Direction Acc.') and the appropriateness of the predicted expansion amount ('Proportion Suitability') were assessed via manual verification.

The results highlight the limitations of simpler strategies. The 'Bbox Only' baseline achieves a relatively high Direction Accuracy, as direct edge contact is a strong indicator, but it cannot predict any expansion proportion (N/A) and fails when objects are occluded before reaching the edge. Relying solely on 'Agent Reasoning (Image Only)' yields a lower Direction Accuracy, suggesting difficulty in precisely judging edge interactions without a geometric cue, although it produces reasonably suitable proportions when it does predict expansion.

Our proposed hybrid approach ('Ours (Image + Bbox Prior)') demonstrates superior performance, achieving the highest Direction Accuracy and the best Proportion Suitability. This indicates that providing the bounding box contact as a geometric prior significantly aids the reasoning agent in correctly identifying truncated edges, while the subsequent reasoning leveraging both image context and this prior leads to the most appropriate estimation of the expansion amount. This study validates the effectiveness of combining geometric checks with contextual reasoning for robust boundary expansion analysis in our framework.

\begin{table}[ht] 
\caption{Ablation study on occlusion relationship prediction. Accuracy reflects overall correctness. Hallucination Rate is the False Positive Rate (predicting occlusion when none exists).}
    \centering
    \resizebox{\columnwidth}{!}{%
    \begin{tabular}{lccc}
        \toprule
        Reasoning Approach & Accuracy (\%) ($\uparrow$) & Hallucination Rate (\%) ($\downarrow$) & \\
        \midrule
        Single Agent (GPT-4o only) & 87.2 & 10.5 &  \\ 
        \midrule 
        \multicolumn{4}{l}{\textit{Ours (Collaborative Multi-Agent Framework)}} \\ 
         \quad w/ GPT-4o        & \textbf{92.5} & \textbf{5.1} \\ 
         \quad w/ Gemini 2.5 Pro \cite{Ref_Gemini} & 91.5 & 8.2 \\ 
        \bottomrule
    \end{tabular}
    }
    \label{tab:occlusion_reasoning_ablation}
\end{table}

\begin{table}[ht] 
\caption{Ablation study on boundary expansion prediction. Direction Accuracy reflects correct identification of edges needing expansion. Proportion Suitability reflects the rate at which the predicted expansion amount is deemed appropriate.}
    \centering
    
    \resizebox{\columnwidth}{!}{%
    \begin{tabular}{lcc} 
        \toprule
        Boundary Analysis Strategy        & Direction Acc. (\%) ($\uparrow$) & Proportion Suitability (\%) ($\uparrow$) \\ 
        \midrule
        Baseline: Bbox Touching Edge Only & 88.61                            & N/A* \\ 
        Agent Reasoning (Image Only)      & 80.15                            & 90.79 \\ 
        \textbf{Ours (Image + Bbox Prior)}& \textbf{90.79}                   & \textbf{92.02} \\ 
        \bottomrule
    \end{tabular}
    }
    \label{tab:ablation_boundary_detail} %
\end{table}

\section{Dataset}
\subsection{Open-World Benchmark Components}
For benchmarking our method's performance on diverse, open-world scenarios and facilitating comparison with recent state-of-the-art methods, we adopt the dataset construction principles and source datasets utilized by Ao et al. \cite{ao2024open}. However, due to significant challenges and instability encountered when attempting to download the large-scale LAION dataset component at the time of our experiments, we excluded LAION from our evaluation set based on this benchmark.

\subsection{Custom Diverse Dataset}
To further evaluate robustness and generalization, particularly on scenarios potentially underrepresented in standard benchmarks (e.g., specific types of real-world clutter, varied lighting), we constructed a complementary custom dataset. This dataset was curated specifically by us to increase diversity and comprises:

Everyday Photographs: A set of 50 images captured by the authors using mobile devices in various real-world indoor and outdoor settings (e.g., homes, offices, street views). These feature naturalistic lighting and common object occlusions encountered in daily life. Target objects were manually selected. 

Royalty-Free Internet Images: A collection of 200 high-resolution images downloaded from public domain or royalty-free sources (e.g., Unsplash, Pexels). These were gathered using keywords related to common objects and occlusion, followed by manual filtering to select high-quality examples with clear partial occlusions suitable for the task.

Evaluating on both the established benchmark components (excluding LAION) and our diverse custom dataset provides a thorough testbed for assessing the capabilities and limitations of amodal completion methods across a wide range of conditions.

\end{document}